\documentclass[a4paper,english]{rnti}

\usepackage[T1]{fontenc}
\usepackage[utf8]{inputenc}
\usepackage[]{textgreek}
\usepackage{multirow}
\usepackage{makecell, caption, booktabs, multirow, siunitx}
\usepackage{amsmath}

\newcommand{\otoprule}{\midrule[\heavyrulewidth]}
\usepackage{lipsum, fancyvrb}

\usepackage{url}

\usepackage{graphicx}

\titrecourt{Experiments on transfer learning architectures for biomedical relation extraction}

\nomcourt{W. Hafiane et al.}

\titre{Experiments on transfer learning architectures for biomedical relation extraction}

\auteur{Walid Hafiane\affil{1},
         Joël Legrand\affil{1},
         Yannick Toussaint\affil{1},
         Adrien Coulet\affil{1}\affilsep\affil{2}
         }

\affiliation{
     \affil{1}Loria, CNRS, Inria, Université de Lorraine, 
     54506 Vandœuvre-lès-Nancy, France\\
          \http{http://www.loria.fr/}\\
     \affil{2}Inria Paris, Inserm UMR1138, Université de Paris, France\\
           \{first name.last name\} @loria.fr\\
  }

\resume{
Relation extraction (RE) consists in identifying and structuring automatically relations of interest from texts. Recently, BERT improved the top performances for several NLP tasks, including RE. However, the best way to use BERT, within a machine learning architecture, and within a transfer learning strategy is still an open question since it is highly dependent on each specific task and domain. Here, we explore various BERT-based architectures and transfer learning strategies (i.e., \emph{frozen} or \emph{fine-tuned}) for the task of biomedical RE on two corpora. Among tested architectures and strategies, our *BERT-segMCNN with fine-tuning reaches performances higher than the state-of-the-art on the two corpora (1.73 \%  and 32.77 \% absolute improvement on ChemProt and PGxCorpus corpora respectively). More generally, our experiments illustrate the expected interest of fine-tuning with BERT, but also the unexplored advantage of using structural information (with sentence segmentation), in addition to the context classically leveraged by BERT.
}

 \summary{
 \vspace{-0.2cm}
 L'extraction de relations (ER) consiste à identifier et à structurer automatiquement des relations à partir de textes. Récemment, BERT a permis d'améliorer les performances de plusieurs tâches de NLP, dont l'ER. Cependant, la meilleure façon d'utiliser BERT, dans une architecture d'apprentissage automatique avec une stratégie d'apprentissage par transfert reste une question ouverte, car elle dépend de la tâche et du domaine d'application. Dans ce travail, nous explorons diverses architectures d'ER qui s'appuient sur BERT et plusieurs stratégies de transfert (i.e., \emph{frozen} ou \emph{fine-tuning}) sur deux corpus biomédicaux. Parmi les architectures et les stratégies de transfert testées, *BERT-segMCNN avec fine-tuning atteint des performances supérieures à l'état de l'art sur deux corpus (amélioration absolu de 1,73\% et 32,77\% sur ChemProt et PGxCorpus respectivement). Plus généralement, nos expériences illustrent l'intérêt attendu d'utiliser le fine-tuning avec BERT, ainsi que l'utilité de l'ajout de l'information structurelle en considérant la segmentation des phrases, en plus du contexte exploité par BERT. 
 }

\begin{document}

%
\section{Introduction}

The biomedical literature is continuously growing in volume, which makes natural language processing (NLP) an essential tool for leveraging domain knowledge it expresses.
In this context, the NLP task of relation extraction (RE) plays an important role in many applications, such as semantic understanding, query answering or knowledge summarization. 
Indeed, RE allows to extract and structure elements of knowledge from natural language texts by identifying and typing the relationships that may be mentioned between named entities \citep{Pawar2017}. The resulting relations can be normalized and assembled in the form of a knowledge graph (KG), which summarizes a domain of knowledge and may serve as a structured intermediate for subsequent tasks of knowledge discovery or knowledge comparison \citep{monnin:hal-02103899}. 
In this work, we explore the task of RE, in the biomedical domain, and from a deep learning perspective.\par
Deep neural networks have achieved significant success for the task of RE \citep{Kumar2017}, and more generally for NLP tasks \citep{Collobert11}. In particular, convolutional and recurrent neural networks (CNN and RNN) have been successfully used for the task of biomedical RE. 
Inspired from the vision domain, multi-channel CNN (MCNN) have been used in NLP domain for RE \citep{Quan2016}, because of their ability to extract local features latent in the several embedding vectors. Here, the convolution kernels allow for extracting relevant features in a hierarchical manner. In addition, weight sharing implies a processing time advantage for this architecture. 
In contrast, RNN is an architecture that can capture the semantic features in the context. The recurrence allows for storing representations of the input sequence in the form of an activation. The last hidden state summarizes the context of the whole input, and  thus is considered as a semantic representation at the sentence level. Bidirectional Long Short-Term Memory (BiLSTM) \citep{Hochreiter1997} is a RNN broadly used in NLP mainly for its ability to capture relatively short past and future context. \par
Indeed, previously mentioned architectures fail at capturing long contexts. This led to the adoption of attention mechanism in order to capture long-distance relations \citep{Chen2020}. 
Leveraging this idea, BERT (Bidirectional Encoder Representations from Transformers) combines encoders and attention mechanisms, and then 
advances the level of performances in many NLP tasks including RE \citep{Devlin2018, Shi2019}. 
BERT principle, such as Elmo (Embeddings from Language Models), is based on transfer learning \citep{Peters2018}. It consists of pre-training a language models on large corpora with unsupervised tasks, which provides a general language understanding, subsequently used in downstream NLP tasks.


For instance, BERT models are strong on learning semantic features due to transformer attention architecture, but they are less capable to capture the structural information. Indeed, no explicit order is imposed between the representation vectors in its different blocks, particularly in the attention sublayer. This leads to a weak contribution of the structural information to BERT output. The unique source of this information is positions embedding included in the inputs, this embedding structural information propagates through the 12 transformer blocks of BERT, which results in a poor representation of this information. In contrast, \cite{Li2019} demonstrate that using sentence segmentation as pre-processing may improve an RNN architecture by providing structural information. \cite{Chen2020} are  exploring this pre-processing with a CNN architecture. 

Our work explores two transfer learning strategies to enhance BERT variants (denoted *BERT) performances for the task of biomedical RE. We propose BERT-based architectures for ``frozen'' and ``fine-tuning'' transfer learning strategies, in an attempt to extract more relevant features, out of BERT representation vectors. Moreover, we explore strengthening BERT with structural information using sentence segmentation as post-processing, which is -up to our knowledge- an original approach. 
We tested on two benchmarks biomedical corpora ChemProt \citep{Kringelum2016} and PGxCorpus \citep{Legrand2020} .


This article is structured as follows: Section 2 provides elements of background on our learning task (biomedical RE), on BERT architecture and transfer learning strategies. Section 3 details the BERT-based architectures and transfer learning strategies we implemented. Section 4 exposes experiments and their results. Section 5 discusses our results and concludes.

\section{Background}

\subsection{Relation Extraction Task}

\begin{figure}[h]
\vspace{-0.5cm}
\begin{center}
 \includegraphics[width=12cm]{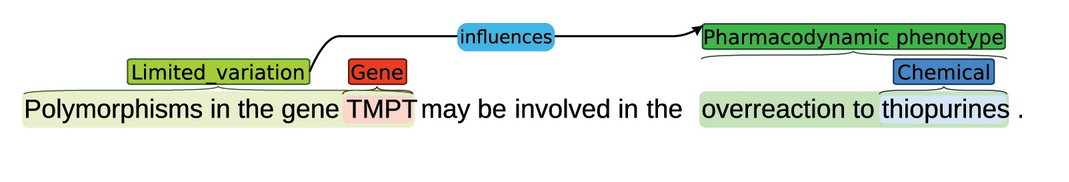}
 \caption{Example of sentence with four named entities and a relation between two of them. This relation is an extract from PGxCorpus\citep{Legrand2020}.}
 \label{fig:relation_exemple}
\end{center}
\end{figure}

Relation extraction (RE) aims at identifying, in unstructured text, all the instances of a predefined set of types of relations between identified entities.
As a result, relations are composed of two or more named entities and a label that type the association between them. 
Figure \ref{fig:relation_exemple} provides an example of relation of the type ``influences'' between two entities (of type Limited\_variation and Pharmacodynamic\_phenotype). We are considering here binary RE that can be seen as a classification task by computing a score for each possible relation type, given a sentence and two identified entities.



\subsection{Two Example Corpora with Annotated Relations}

We experimented with two manually annotated corpora which focus on slightly different kind of biomedical relationships. 

\emph{ChemProt} is a corpora of relationships between proteins and chemicals, with 10,031 examples, distributed in train, validation, and test sets \citep{Kringelum2016}. Annotated relations are of 13 different types of protein-molecule relations. Table (\ref{Dist_ChemProt}) shows the distribution of examples over relation types in the train set. ChemProt is particularly unbalanced since the most frequent relation type in the train set represents 39.4\%, while the least frequent type represents only 0.1\%. For instance, ``AGONIST INHIBITOR'' has only 4 examples in the train set.

\emph{PGxCorpus} is a corpora of pharmacogenomic relationships between genomic factors and drugs or drug responses \citep{Legrand2020}. It is composed of 2,875 examples, distributed over 7 different types of relations (isAssociatedWith, influences, causes, decreases, increases, treats, isEquivalentTo). The percentage and number of examples in each type of relationship are presented in table(\ref{Dist_PGxCorpus}). The distribution of relationships is unbalanced: the most frequent is ``influences'' (32.6\%), while ``causes'' is the least frequent  (5.8\%).

The two corpora were chosen for distinct reasons: ChemProt is a well established corpus that served as Benchmark in many previous work on transfer learning, enabling us to compare our performances with state-of-the-art approaches; PGxCorpus is a novel corpus for the domain of Pharmacogenomics, which offers opportunities for studying the extraction of n-ary and/or nested relationships. 

\begin{table}[h]
	\begin{center}
	    \vspace{-0.45cm}
		\begin{tabular}{ccc}
			\hline\hline
			Range & Class & Size \\
			\hline
			$>$1000 & INHIBITOR & 1642(39.4\%) \\ 
			$>$400 & I.-DOWNREGULATOR / SUBSTRATE& 487(11.7\%) / 480(11.5\%)\\ 
			$>$300 & I.-UPREGULATOR / ACTIVATOR &  387(9.3\%) / 323(7.7\%)\\	
			$>$200 & ANTAGONIST / PRODUCT-OF& 235(5.6\%) / 233(5.6\%)\\	
			$>$100 & AGONIST / DOWNREGULATOR & 156(3.7\%) / 131(3.1\%) \\
			$>$10 & UPREGULATOR / SUBSTRATE-P.& 67(1.6\%) / 14(0.3\%)\\
			$\leq$10 & AGONIST-A. / AGONIST-I. & 10(0.2\%) / 4(0.1\%)\\
			\hline
		\end{tabular}
		\caption{Distribution of relationships by type in ChemProt.} \label{Dist_ChemProt}
	\end{center}
\end{table}

\begin{table}[h]
    \vspace{-0.5cm}
	\begin{center}
		\begin{tabular}{ccc}
			\hline\hline
			Range & Class & Size \\
			\hline
			$>$400 & Influences / IsAssociatedWith  & 937(32.6\%) / 733(25.5\%) \\
			$>$250 & IsEquivalentTo / Decreases  &  293(10.2\%) / 263(9.1\%) \\
			$>$200 & Increases / Treats & 243(8.5\%) / 238(8.3\%) \\
			$\leq$200 & Causes & 168(5.8\%) \\
			\hline
		\end{tabular}
		\caption{Distribution of relationships by type in PGxCorpus.} \label{Dist_PGxCorpus}
	\end{center}
\end{table}

\subsection{BERT}

\paragraph{}
BERT \citep{Devlin2018} is the first encoder architecture that captures past and future context simultaneously, unlike OpenAI GPT (Improving Language Understanding by Generative Pre-Training) \citep{Radford2018}, which only captures past or Elmo \citep{Peters2018} which uses a concatenation of the independently captured past and future contexts. BERT encoder is made up of a stack of 12 identical blocks, where each block is composed of two sublayers. The first one is a multi-head self-attention mechanism, and the second is a fully connected feedforward neural network. Residual connections are used around each of the two sublayers, followed by layer normalization applied after each sublayer. The transformer is pre-trained on two unsupervised tasks (i.e., prediction of masked words, prediction of the next sentence) on large corpora (i.e., Wikipedia of 2.5 billion words, BookCorpus of 800 million words). The aim of the pre-training is to acquire a general language understanding, with the idea of transferring this ability to downstream tasks.

\subsection{Transfer Learning}

According to \citep{JialinPan2010} transfer learning is divided into two main categories: \emph{inductive} and \emph{transductive transfer}. 
Inductive transfer consists of transferring information between different tasks, usually within the same domain. 
Whereas transductive transfer consists in transferring information from one domain to another, within the same task.

Following this transductive approach, the initial BERT model has been used to develop models particularly adapted to specific domain and languages, such as those of biomedicine. It results from this transfer several variants of BERT that use the pre-trained BERT model and apply a domain transfer in order to make BERT more efficient in the target domain. 
In particular, BioBERT initializes its weights with BERT pre-trained model and then continue the training on a large set of biomedical texts (about 18 billion words) \citep{Lee2020}. BERT and BioBERT performances have already been compared on the task of RE on ChemProt corpus, showing that BioBERT was outperforming.
SciBERT is another BERT variant that follows the same approach, but uses 3.17 billion words of mixed-scientific texts \citep{Beltagy2019}, out of which 82\% comes under the biomedical domain. SciBERT is the state-of-the-art model on the task of RE on ChemProt.

Following the inductive approach, BERT pre-training model and its ability to understand, or represent language, is typically reused and enriched (e.g., with an additional layer of neurons) for different tasks such as named entity recognition, or RE. 
In this context, we explore both \emph{frozen} and \emph{fine-tuning} strategies. The \emph{frozen} strategy consists in reusing layers of neurons from a previously trained model, and ``freeze'' them, i.e., avoiding to update its parameters during future training steps. Next, one usually adds layers on top of the frozen layers so these new layers will consider output of frozen layers as input features for the final learning task. 
In contrast, the \emph{fine-tuning} strategy also reuses layers from a previously trained model, plus additional layers on top, but allows to tune the parameters of the whole model in regards with the new train set.
In the frozen strategy, the fact that the gradient back-propagation applies only to the added neurons presents an advantage in terms of processing time, whereas fine-tuning may obtain increased performance by adapting the pre-trained parameters to the new train set.

In this work we explore both transductive and inductive transfer learning for the tasks of biomedical RE, by experimenting not only various BERT variants, but also both frozen and fine-tuning strategies.

\vspace{-0.3cm}
\section{Method}

\subsection{Architectures for Relation Extraction}

\subsubsection{State-of-the-art Architectures}


\paragraph{*BERT+BiLSTM}
In the frozen strategy, our baseline is the frozen architecture reported for RE in the SciBERT article, which consists in an RNN-based model on top of a BERT-like model\citep{Beltagy2019}. For simplicity, we denoted BERT-like model, i.e., either BERT or BioBERT or SciBERT with the *BERT. In this architecture, *BERT is used as a pre-trained contextualized word embeddings. The added classifier is composed of a 2-layer BiLSTM of size 200, followed by a multilayer perceptron applied on the concatenated first and last BiLSTM vectors. This RNN part is supposed to extract the contextual information from embeddings vectors sequence, in order to feed the multilayer perceptron classifier. For short, and depending on the frozen pre-trained model, we denote this baseline architecture by BERT+BiLSTM, BioBERT+BiLSTM and SciBERT+BiLSTM, or more generally *BERT+BiLSTM.

\paragraph{*BERT+MLP}
In the fine-tuning strategy, our baseline is the architecture reported for RE in BERT, BioBERT and SciBERT. It consists in a simple multilayer perception (MLP), composed of a single fully connected hidden layer, as an extension on the top of BERT pre-trained models \citep{Devlin2018, Lee2020, Beltagy2019}. In the following, we denote these architectures with BERT+MLP, BioBERT+MLP and SciBERT+MLP. 

In the original SciBERT article, RE has been evaluated on ChemProt following both the frozen (SciBERT+BiLSTM) and fine-tuning (SciBERT+MLP) strategies. We reproduced voluntarily their results for further comparisons, with other architectures or corpus (i.e., PGxCorpus).

\subsubsection{Proposed Architectures}

\begin{figure}[h]
\vspace{-0.5cm}
	\begin{center}
		\includegraphics[angle=-90,scale=0.5]{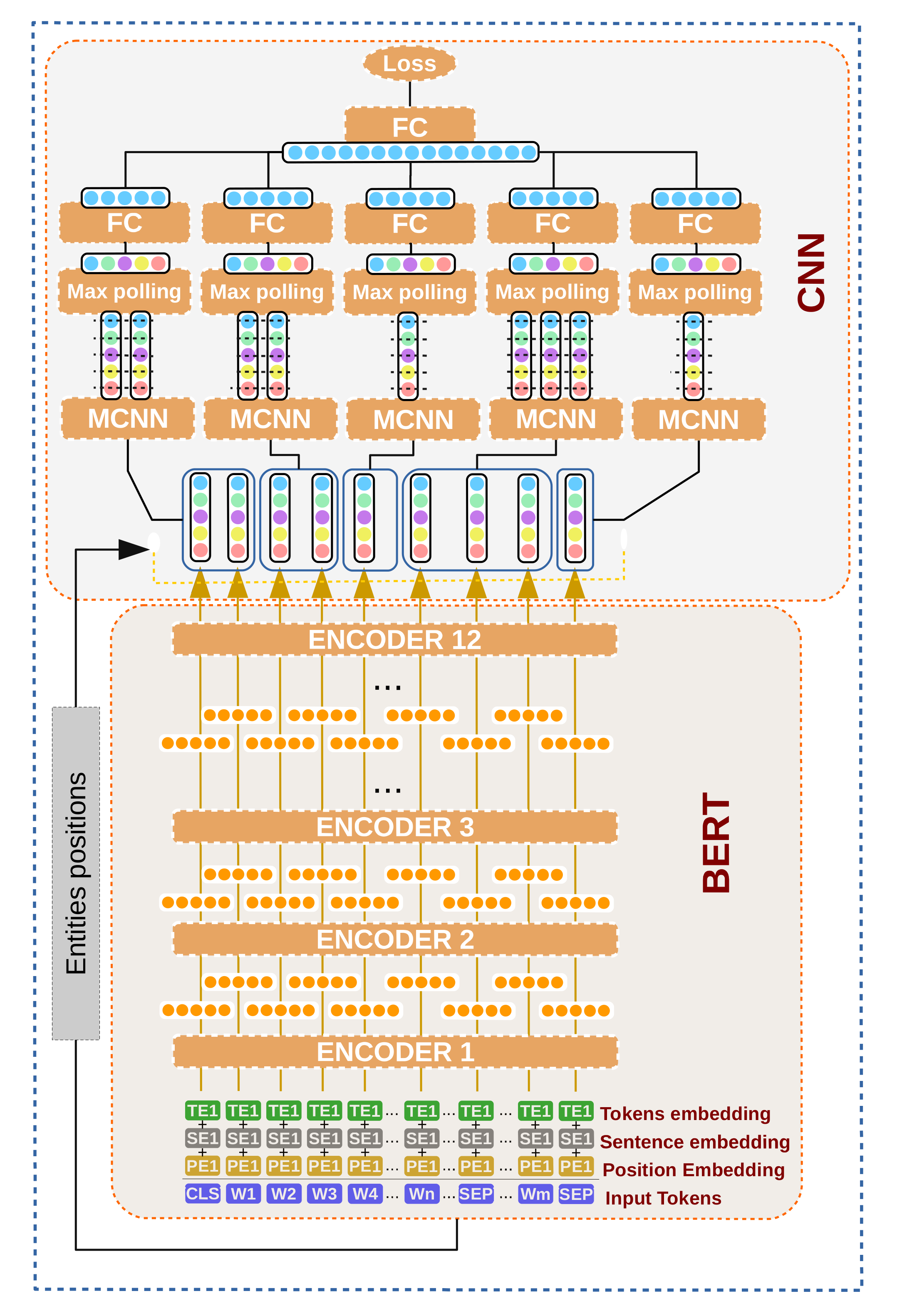}
		\caption{Overview of our BERT-segMCNN architecture.} 
		\label{fig:BERT+segMCNN}
	\end{center}
\end{figure}

\paragraph{*BERT+MCNN}
The first proposed architecture is the extension of BERT with a MCNN (denoted generally *BERT+MCNN). We experiment this combination, with the hope that MCNN extracts local information from the representation vectors computed by BERT.  Indeed BERT has the ability to extract long distance contextual information using multi-headed attention, whereas local contextual information captured by the MCNN may provide additional discriminated features. The fact that the order between the different dimensions is not explicit in the representation vectors lead us to consider them as independent features and handling each dimension as a different embeddings. Accordingly, each dimension represents a channel of the MCNN. This architecture is followed by a max-pooling to obtain a sentence representation vector, which in turn fed a multilayer perceptron. MCNN contains only few parameters (due to parameters sharing), no recurrence and relatively few layers, which globally tends to reduce the internal covariate shift effect. This let us envisage computing fine-tuning with this architecture (in addition to the frozen, that is less computationally expensive). As a results, we experiment this architecture with both frozen and fine-tuning strategies.

\paragraph{*BERT+BiLSTM-MCNN}
The second type of proposed architectures extends BERT with a BiLSTM and a MCNN. The exploration of this architecture is motivated by the fact that BiLSTM may capture semantic features in BERT representation vectors and that MCNN may capture the local information from the same vectors. To explore how these two kind of features may be combined we proposed two *BERT+BiLSTM-MCNN architectures: one linear (*BERT+BiLSTM-MCNN L.) where BiLSTM outputs input MCNN, before a maxpooling and a multilayer perceptron for classification ; one parallel (*BERT+BiLSTM-MCNN P.) where both MCNN and BLSTM inputs are BERT representation vectors. The output of MCNN and BiLSTM are then concatenated to feed the multilayer perceptron for classification. These architectures are experimented only in the frozen strategy, mainly because in the fine-tuning training the depth of BERT, plus BiLSTM-MCNN extension makes the computation very heavy and may lead to a vanishing gradient problem.

\paragraph{*BERT+segMCNN}
The third proposed architecture extends BERT with five MCNN, each of them taking as an input a different partition (or segment) of BERT representation vectors.  
We propose this architecture because classically, no order is imposed in the transformer between representation vectors, in particular in the attention layer. As a result, the structural information may be limited in the output representations vectors. In order to strengthen *BERT architecture with this information, we proposed five MCNN, in parallel, as illustrated in Figure (\ref{fig:BERT+segMCNN}). The segmentation of the representation vectors used is based on the location of the entities in the sentence. Representation vectors are split according to their belonging to one of the following five partitions: before the first occurring entity, first entity, between the first entity and second entity, second entity, and after the second entity. In order to carry out this post-processing, entities positions are reported from the input of the whole architecture to the input of the segMCNN block, as shown Figure(\ref{fig:BERT+segMCNN}).
Max polling and fully connected (FC) layer are applied after each MCNN, which provides five vectors, each one representing one segment. This architecture ends with an MLP used for a classification of the sentence representation vector, obtained by the concatenation of the five segment vectors.

\subsection{Experimental Setup}

We setup 4 batches of experiments, each of them associated with one target corpus (Chem-\\Prot or PGxCorpus) and one transfer strategy (frozen or fine-tuning). In each batch, we tested for comparison both state-of-the-art architectures and the original ones we proposed. 

\subsubsection{Evaluation Metrics}
Depending on the experiments we used as evaluation metrics either the \emph{macro} or the \emph{micro averaged F-measure} (noted F-macro and F-micro respectively). F-measure combines precision and recall by computing their harmonic mean.
When reporting a unique F-measure for a multi-class classification task, there is various ways to average the F-measure: either by considering independently each example i.e., averaging on all examples treating them equally regardless of their class (for F-micro); or by a simple arithmetic average on classes treating all classes equally regardless of their size (for F-macro). 
Classically, F-micro is preferred in highly unbalanced settings. In our case, we preferred F-macro in experiments with PGxCorpus, and F-micro with ChemProt because the latest includes rare types of relationships. We note that F-micro is computationally equivalent to the accuracy metric in the case of disjoint classes, which is the case in this work. In both cases, F-measure by class (i.e., by type of relationship) are available in an online supplementary material (see result section).


\subsubsection{Experimental Settings}

Models are trained to minimize the cross-entropy function. 
With ChemProt corpora, our models are trained on the train set, tested on the test, and the validation set is used for model selection and hyperparameter tuning. We reused the hyperparameters tuned on ChemProt for PGxCorpus experiments.
For PGxCorpus no train, validation and test sets are predefined. Thus we adopted a 10-fold cross-validation strategy to define them randomly several times.
Results of hyperparameter tuning provided us with various optimal setting for each pair architecture-transfer strategy. Accordingly we report here the sets of parameters values we used. 
For CNN, the convolution filter sizes are (3, 5, 7) and the number of filters of (3, 6). For BiLSTM we used two-layers of size 200, and for MLP one hidden layer of size (64, 100). We train our models with a batch size of 32 and use two regularization types: a dropout of (0.1, 0.25, 0.5) and a L2-regularization of (0, 0.01). We optimize the loss function using AdamWithDecay with a decay of (0, 0.01) and an initial learning rate of 0.001 for the frozen transfer strategy, and ($3.10^{-5}$ ,$5.10^{-5}$ ,$10^{-5}$) for the fine-tuning strategy. Numbers of epochs used are (30, 65) for the frozen strategy, and (5, 8) for the fine-tuning strategy. Added weights are initialized by a normal distribution ($\mu=0, \sigma=0.02$). The learning procedure is initialized with different random weights (about 100 times) and we report the mean performances. We reported the standard deviation to analyze the stability of our models.

Experiments has been developed with Python and PyTorch libraries, with the version BERT-base-uncased of BERT, v1.1 of BioBERT-uncased and SciBERT-SciVOCAB-uncased of SciBERT. 
\footnote{The programmatic code of our experiments is available at \url{https://github.com/hafianewalid/Transfer-Learning-Architectures-for-Biomedical-Relation-Extraction}}

\vspace{-0.3cm}
\section{Experimental Results}

\subsection{Frozen Strategy}
\begin{table}[h]
  \renewcommand{\arraystretch}{1.3}
  
  \vspace{-0.5cm}
  \centering
  \begin{tabular}{lccc}
  \hline\hline
    & {\thead{Architecture}} & {\thead{F-micro}} & {\thead{$\sigma$}} \\
    \otoprule
    \multirowcell{1}{\citep{Beltagy2019}} & SciBERT+BiLSTM & 75.03 & – \\
    \hline\hline
    \multirowcell{4}{BERT}             & + BiLSTM & 64.41 & 2.42 \\
                                       & + MCNN & 68.26 & 1.54 \\
                                       & + BiLSTM-MCNN L. & 75.35 & 1.02 \\
                                       & + BiLSTM-MCNN P. & 62.07 & 1.33 \\
    \hline\hline
    \multirowcell{4}{BioBERT}      & + BiLSTM & 72.86 & 1.36 \\
                                   & + MCNN & 77.57 & 0.70 \\
                                   & + BiLSTM-MCNN L. & \textbf{80.08} & 0.80 \\
                                   & + BiLSTM-MCNN P. & 74.85 & 0.96 \\
    \hline\hline
    \multirowcell{3}{SciBERT}      & + BiLSTM (SOTA Reproduction) & 75.10 & 1.00 \\    
                                   & + MCNN & 77.85 & \textbf{0.68} \\
                                   & + BiLSTM-MCNN L. & 79.24 & 0.76 \\
                                   & + BiLSTM-MCNN P. & 70.45 & 1.18 \\
                                   
    \hline                               
  \end{tabular}
  \caption{Evaluation of performances of various transfer learning architectures on ChemProt, using the frozen strategy.}
  \label{tab:ChemProt-frozen}
  \vspace{-0.3cm}
\end{table}

\begin{table}[h]
  \renewcommand{\arraystretch}{1.3}
  \vspace{-0.5cm}
  \centering
  \begin{tabular}{lccccc}
  \hline\hline
    & {\thead{Architecture}} & {\thead{Precision}} & {\thead{Recall}} & {\thead{F-macro}} & {\thead{$\sigma$}} \\
    \otoprule
    \multirowcell{1}{\citep{Legrand2020}}  & MCNN & – & – & 45.67 & 4.51\\
    \hline\hline
    \multirowcell{4}{BERT}       & + BiLSTM & 54.29 & 54.82 & 54.29 & 0.84\\
                                 & + MCNN & 57.64 & 53.84 & 53.73 & 1.33 \\
                                 & + BiLSTM-MCNN L. & 71.85 & 70.47 & 70.63 & 1.38 \\
                                 & + BiLSTM-MCNN P. & 54.48 & 54.37 & 53.99 & \textbf{0.28} \\
    \hline\hline
    \multirowcell{4}{BioBERT} & + BiLSTM & 57.16 & 57.32 & 56.93 & 0.89 \\
                                   & + MCNN & 69.52 & 66.25 & 67.02 & 2.43 \\
                                   & + BiLSTM-MCNN L. & \textbf{73.05} & 71.81 & 72.09 & 1.71 \\
                                   & + BiLSTM-MCNN P. & 67.46 & 64.64 & 65.39 & 0.57 \\
    \hline\hline
    \multirowcell{4}{SciBERT}
                                   & + BiLSTM & 62.60 & 62.26 & 62.18 & 1.18 \\
                                   & + MCNN & 69.59 & 66.57 & 67.35 & 1.74 \\
                                   & + BiLSTM-MCNN L. & 72.65 & \textbf{72.58} & \textbf{72.38} & 1.04 \\
                                   & + BiLSTM-MCNN P. & 61.54 & 61.11 & 61.00 & 1.45 \\
                                   
    \hline                               
  \end{tabular}
  \caption{Evaluation of performances of various transfer learning architectures on PGxCorpus, using the frozen strategy.}
  \label{tab:PGxCorpus-frozen}
  \vspace{-0.3cm}
\end{table}

\paragraph{With ChemProt corpus}
Results obtained with the frozen strategy and ChemProt corpus are presented in Table \ref{tab:ChemProt-frozen}. The first line of Table \ref{tab:ChemProt-frozen} reports state-of-the-art (SOTA) results from \citep{Beltagy2019}, which we reproduced and reported on the 10$^{th}$ line of the table (denoted SOTA reproduction). It drifts of 0.07\% F-micro, within a standard deviation of 1. 

Regardless the variant of BERT that we used, *BERT+BiLSTM-MCNN L. models produces the most performing models on ChemProt corpus. BioBERT+BiLSTM-MCNN achieves the higher performance with 80.08 \% F-micro.

\paragraph{With PGxCorpus}
Results obtained with the frozen strategy and PGxCorpus are presented in Table \ref{tab:PGxCorpus-frozen}. The first line reports SOTA results from \citep{Legrand2020}, obtained with a simple MCNN, without any BERT pre-trained model.

As expected, we observe that all our (BERT-based) architectures exceed this baseline, and are more stable. In particular, we note that independently of the variant of BERT used, the architecture *BERT-BiLSTM-MCNN L. produces the most performing models on PGxCorpus, similarly to what we observed with ChemProt.
We also note that, this time SciBERT+BiLST-\\M-MCNN L. overpass (slightly) BioBERT version of the same architecture.  



\vspace{-0.2cm}
\subsection{Fine-tuning Strategy}

\begin{table}[h]
  \renewcommand{\arraystretch}{1.3}
  \vspace{-0.5cm}
  \centering
  \begin{tabular}{lccc}
  \hline\hline
    & {\thead{Architecture}} & {\thead{F-micro}} & {\thead{$\sigma$}} \\
    \otoprule
    \multirowcell{1}{\citep{Beltagy2019}} & SciBERT+MLP & 83.64 & – \\
    \hline\hline
    \multirowcell{3}{BERT}      & + MLP & 79.28 & 1.21 \\
                                & + MCNN & 72.18 & 1.73 \\
                                & + segMCNN & 81.43 & 0.91 \\
    \hline\hline
    \multirowcell{3}{BioBERT}   & + MLP & 82.28 & 3.27 \\
                                & + MCNN & 83.90 & 1.11 \\
                                & + segMCNN & \textbf{85.37} & \textbf{0.67} \\
    \hline\hline
    \multirowcell{2}{SciBERT}   & + MLP (SOTA Reproduction) & 82.44 & 1.47\\ 
                                & + MCNN & 82.98 & 0.96 \\
                                & + segMCNN & 84.77 & \textbf{0.67} \\
                                
    \hline   
  \end{tabular}
  \caption{Evaluation of performances of various transfer learning architectures on ChemProt, using the fine-tuning strategy.}
  \label{tab:ChemProt-fine-tuning}
\end{table} 

\begin{table}[!hbt] 
  \renewcommand{\arraystretch}{1.3}
  \vspace{-0.4cm}
  \centering
  \begin{tabular}{lcccccc}
  \hline\hline
    & {\thead{Architecture}} & {\thead{Precision}} & {\thead{Recall}} & {\thead{F-macro}} & {\thead{$\sigma$}} \\
    \otoprule
    \multirowcell{1}{\citep{Legrand2020}}  & MCNN & – & – & 45.67 & 4.51\\
    \hline\hline
    \multirowcell{3}{BERT}         & + MLP & 70.80 & 69.70 & 69.88 & 4.03 \\
                                   & + MCNN & 71.73 & 73.39 & 72.22 & \textbf{0.98} \\
                                   & + segMCNN & 74.31 & 74.53 & 74.17 & 1.08 \\
    \hline\hline
    \multirowcell{3}{BioBERT}      & + MLP & 73.49 & 68.84 & 70.47 & 5.87 \\
                                   & + MCNN & 74.40 & 71.16 & 72.23 & 4.30 \\
                                   & + segMCNN & 77.38 & 77.24 & 77.00 & 2.56 \\
    \hline\hline
    \multirowcell{3}{SciBERT}      & + MLP & 75.61 & 75.44 & 75.32 & 2.11 \\
                                   & + MCNN & 75.88 & 76.08 & 75.70 & 1.04 \\
                                   & + segMCNN & \textbf{78.15} & \textbf{79.17} & \textbf{78.44} & 1.07 \\
    \hline
  \end{tabular}
  
  \caption{Evaluation of performances of various transfer learning architectures on PGxCorpus, using the fine-tuning strategy.}
  \label{tab:PGxCorpus-fine-tuning}
\end{table}

\paragraph{With ChemProt corpus}
Results obtained with the fine-tuning strategy and ChemProt corpus are presented in Table \ref{tab:ChemProt-fine-tuning}. The first line of Table \ref{tab:ChemProt-fine-tuning} reports SOTA results from \citep{Beltagy2019}, which we reproduced and reported on the 8$^{th}$ line of the table (denoted SOTA reproduction). It drifts of 1.2\% F-micro, within a standard deviation of 1.47. 


We observe that for each variant of BERT, models obtained with *BERT+segMCNN exceeds the performances of the others. In particular we note that BioBERT+segMCNN let us reach a new state-of-the-art performance (85.37\%, absolute improvement of 1.73\%). We also observe that the addition of our segmentation approaches with our 5 MCNN in parallel (segMCNN)  surpasses the SOTA results with both BioBERT and SciBERT as a base.


\vspace{-0.1cm}
\paragraph{With PGxCorpus}
Results obtained with the fine-tuning strategy and PGxCorpus are presented in Table \ref{tab:PGxCorpus-fine-tuning}. The first line reports SOTA results from \citep{Legrand2020}, obtained with a simple MCNN, without any BERT pre-trained model.

As expected, we observe  that all our (BERT-based) architectures exceed this baseline, and are more stable. 
We observe that for each variant of BERT, *BERT+segMCNN model exceeds the performances of the others. In particular we note that SciBERT+segMCNN let us reach a new state-of-the-art performance on this corpus (78.44\%, absolute improvement of 32.77\%). 
It seems that on this corpus, we observe a first improvement with *BERT+MCNN architecture, and a second one with *BERT+segMCNN. 
We also note that, differently to the results obtained on ChemProt, the best BERT variant is SciBERT (vs. BioBERT with ChemProt).

Experiments with the *BERT+MCNN architectures have been performed in both frozen and fine-tuning strategy, and illustrate that fine-tuning strategy produces much better performances. 

\vspace{-0.5cm}
\section{Discussion and Conclusion}
\vspace{-0.2cm}
We hypothesis that BERT model may be reinforced with elements of structural information through sentence segmentation and by using local information latent in its representation vectors. In this direction, we hope that our empirical approach gives some support to this hypothesis. In addition, our results illustrate the fact that the variant of BERT that one may use impacts final performances. In addition, this impact varies depending on the target corpus and the associated specific task. We observe also, as expected, that fine-tuning strategies, even if computationally demanding, produces generally better performances.

To conclude, we experimented several BERT-based architectures and transfer strategies for the task of RE, with two biomedical corpora.   
We motivated our choice of architectures by the aim of leveraging several types of features: local features extracted by MCNN, contextual features by BiLSTM, and structural features coming from a sentence segmentation approach. We proposed different architectures adapted either to the frozen and fine-tuning strategies (*BERT+BiLSTM-MCNN L. and *BERT+segMCNN, respectively) leading to an improvement of the state-of-the-art performance for our specific tasks of biomedical RE. Even if our empirical contribution is limited to a  subset of variants of BERT and to the specific task of biomedical RE, we think that our architectures (in particular *BERT+BiLSTM-MCNN L. and *BERT+segMCNN) may be suitable for other BERT variants, tasks and domains.


\vspace{-0.45cm}
\bibliographystyle{rnti}
\bibliography{biblio}

\end{document}